\pdfoutput=1
\documentclass[11pt]{article}

\usepackage[preprint]{acl}
\usepackage{booktabs}
\usepackage{times}
\usepackage{latexsym}
\usepackage{tabularx}
\usepackage{enumitem}
\usepackage{expex}
\usepackage{float} 
\usepackage{stfloats} 
\usepackage{amsmath}
\usepackage{placeins}

\usepackage[T1]{fontenc}
\usepackage[utf8]{inputenc}

\usepackage{microtype}

\usepackage{inconsolata}
\usepackage{makecell}
\usepackage{multirow}

\usepackage{graphicx}

\title{\textsc{HebID}: Detecting Social Identities in Hebrew-language Political Text}

\author{First Author \\
  Affiliation / Address line 1 \\
  Affiliation / Address line 2 \\
  Affiliation / Address line 3 \\
  \texttt{email@domain} \\\And
  Second Author \\
  Affiliation / Address line 1 \\
  Affiliation / Address line 2 \\
  Affiliation / Address line 3 \\
  \texttt{email@domain} \\}

\author{
  \textbf{Guy Mor-Lan\textsuperscript{1}},
  \textbf{Naama Rivlin-Angert\textsuperscript{1}},
  \textbf{Yael R. Kaplan\textsuperscript{2}},
  \textbf{Tamir Sheafer\textsuperscript{1}},
  \\
  \textbf{Shaul R. Shenhav\textsuperscript{1}}
\\
  \textsuperscript{1} The Hebrew University of Jerusalem,
  \textsuperscript{2} The Open University of Israel
\\
 \href{mailto:guy.mor@mail.huji.ac.il}{\texttt{guy.mor@mail.huji.ac.il}}
}


\begin{document}
\maketitle

\begin{abstract}

Political language is deeply intertwined with social identities. While social identities are often shaped by specific cultural contexts, existing NLP datasets are predominantly English-centric and focus on coarse-grained identity categories. We introduce \textsc{HebID}, the first multilabel Hebrew corpus for social identity detection. The corpus contains 5,536 sentences from Israeli politicians’ Facebook posts (Dec 2018--Apr 2021), with each sentence manually annotated for twelve nuanced social identities (e.g., Rightist, Ultra-Orthodox, Socially-oriented) selected based on their salience in national survey data. We benchmark multilabel and single-label encoders alongside 2B--9B-parameter decoder LLMs, finding that Hebrew-tuned LLMs provide the best results (macro-$F_1$ = 0.74). We apply our classifier to politicians' Facebook posts and parliamentary speeches, evaluating differences in popularity, temporal trends, clustering patterns, and gender-related variations in identity expression. We utilize identity choices from a national public survey, comparing the identities portrayed in elite discourse with those prioritized by the public. \textsc{HebID} provides a comprehensive foundation for studying social identities in Hebrew and can serve as a model for similar research in other non-English political contexts.\footnote{\url{https://github.com/guymorlan/hebid/}}

\end{abstract}

% achieve a macro-F1 of 0.743, outperforming encoder-only baselines by over six points.

%To demonstrate cross-genre robustness, we evaluate the top model on 500 Knesset speech excerpts, matching social-media performance (macro-F1 = 0.72).

%We apply our classifier across three complementary sources — politician Facebook posts, parliamentary floor speeches, and a national survey of the public, and evaluate differences in the popularity, temporal trends, bundling into clusters, and gender differences in identities. 

\section{Introduction}
Social identities—such as political ideology, religious affiliation, or demographic group membership—are powerful drivers of political behavior and public discourse \citep{tajfel2001integrative}. Yet existing NLP resources for identity detection remain almost entirely English-focused, single-label, and rely on coarse-grained categories (e.g., party or ethnicity). In this work, we introduce \textsc{HebID}, the first publicly released Hebrew dataset for fine-grained, multilabel social identity detection in political text, grounded in both domain expertise and large-scale survey evidence. % We conducted a 14-wave national survey (N=1,769) to identify the twelve social identities most salient to Israeli citizens (e.g., Rightist, Ultra-orthodox, Socially-oriented). 

We frame the task as a sentence-level classification problem: the goal is to determine which, if any, of twelve social identities are being positively expressed or invoked within a given sentence.

The choice of social identities is empirically grounded in survey data. We utilized 12 waves of a national survey (N = 1,769) each of which included questions designed to identify the social identities most salient to Israeli citizens (e.g., Rightist, Ultra-Orthodox, Socially-oriented).
These survey-derived categories were then used to guide the annotation of 5,536 sentences sampled from Facebook posts by Israeli politicians (December 2018--April 2021), ensuring that our labels reflect real-world identity salience.

We benchmark a suite of modern architectures on the Facebook data—(i) multilabel encoder models; (ii) single-label encoder models; and (iii) 2B--9B-parameter decoder LLMs, finding that Hebrew-tuned decoder models (specifically, \textsc{DictaLM2.0}) achieve the highest macro-F$_1$ (0.743), outperforming encoder-only baselines by over six points. We then assess cross-genre generalization by applying our best model to 500 parliamentary speech excerpts, achieving a comparable macro-F$_1$ of 0.72.

Finally, we link three complementary sources in our analysis of how identities behave: politicians’ Facebook posts, parliamentary speeches from the Israeli parliament (Knesset), and survey responses that reflect public identification. We (1) compare identity prevalence and correlations between social media, parliamentary speech, and the public; (2) document how identity discourse surges around election cycles; (3) uncover coherent “bundles” of co-occurring identities; and (4) quantify gender-related variation in identity expression.
%ADD to the previous sentence: demonstrating the potential of HebID to contribute to the analysis of key research questions in the social sciences.

% HebID—comprising survey-anchored taxonomy, annotated text, benchmarks, and linked-source analyses—provides a novel, empirically grounded framework for studying identity rhetoric in non-English political contexts.

%Our contributions are:
%\begin{itemize}
%  \item \textbf{Dataset}: a multilabel Hebrew corpus of 5,536 sentences annotated for 12 social identities, empirically grounded in survey data.
%  \item \textbf{Benchmarks}: comprehensive comparison of encoder-only and seq2seq LLMs, showing the superiority of the latter.
%  \item \textbf{Cross-genre evaluation}: demonstration of model generalization to parliamentary speech.
%  \item \textbf{Sociolinguistic analysis}: novel insights into identity popularity, temporal dynamics, bundling, and gendered expression across multiple sources.
%\end{itemize}
\newpage
Our contributions are:
\begin{itemize}
    \item \textbf{Dataset and survey-grounded methodology}: We introduce a framework for creating identity corpora by linking survey data to text annotation, and release \textsc{HebID}, the resulting Hebrew dataset.
    \item \textbf{Comprehensive benchmarks}: We evaluate a range of encoder-only and decoder LLMs, establishing strong baselines for the task.
    \item \textbf{Cross-genre evaluation}: We demonstrate our best model's generalization to parliamentary speech, confirming its robustness.
    \item \textbf{External validation}: We validate our classifier against an external expert survey (CHES-Israel), showing its alignment with party policy positions.
    \item \textbf{Sociolinguistic analysis}: We use the classifier to reveal novel insights into identity dynamics across social media, parliamentary speeches, and public surveys.
\end{itemize}

Taken together, these contributions offer value on three fronts. For computational social scientists, \textsc{HebID} provides a novel tool for analyzing political discourse. For the NLP community, our work establishes a challenging new benchmark for multilabel classification in a low-resource language. Finally, our survey-grounded annotation framework serves as a methodological template for developing similar culturally-aware resources in other non-English contexts.

\section{Related Work}
Automatic analysis of social identity language in political text intersects several NLP subfields: group reference detection, framing and stance classification, and ideological position inference. However, existing resources remain limited in granularity, language coverage, and multilabel capacity, leaving a clear gap that our Hebrew dataset fills.

\paragraph{Group reference detection.}
Early work extended named-entity recognition to capture social groups as entities. \citet{Zanotto2024} introduce \textsc{GRIT}, annotating Italian news and parliamentary text for spans referring to demographic, national, or partisan groups, and fine-tune BERT to identify them. \citet{LichtSczepanski2024} apply a similar span-labeling approach to British parliamentary debates, quantifying how often politicians mention particular social groups. These studies demonstrate the feasibility of automatic group mention detection, but remain single-label and limited to explicit group mentions.

\paragraph{Framing and stance.}
Beyond mentions, understanding how groups are portrayed is crucial. The \emph{Us vs.\ Them} corpus \citep{HuguetCabot2021} annotates Reddit comments for target group, stance (supportive vs.\ hostile), and emotion, using a multi-task RoBERTa model. \citet{doi:10.1073/pnas.2120510119} study 140 years of U.S.\ congressional speeches on immigration, combining sentiment classification with custom frame lexica to reveal evolving and polarized frames. The Media Frames Corpus \citep{Card2015} provides frame annotations (including cultural identity frames) for news articles. These resources focus on single-label stance or framing, and predominantly English texts.

\paragraph{Ideological position inference.}
Separate but related is the task of inferring latent political ideology. \citet{Thomas2006} use SVMs on floor debate transcripts to classify support/opposition. \citet{Iyyer2014} develop neural networks to predict left/right alignment of speeches. More recent prompt based methods with LLMs \citep{wang-etal-2022-super} achieve zero-shot ideological scoring \citep{le2025positioning}. While these approaches infer broad ideological leanings, they do not detect explicit identity mentions or support multilabel identity categories.

\paragraph{In-group/out-group rhetoric.}
Discourse-level signals such as pronoun clusivity and coded language reveal populist identity appeals. \citet{RehbeinRuppenhofer2022} annotate inclusive vs.\ exclusive uses of “we” in German parliamentary debates and train classifiers to disambiguate referents, highlighting \emph{us-versus-them} framing. \citet{Mendelsohn2023} curate a dogwhistle lexicon and show that pretrained models struggle to detect covert slurs without knowledge grounding. These works emphasize the need for high-precision, context-aware annotation, but do not cover multilabel identity categories across many classes.

\paragraph{Gaps and Our Contributions.}
Current identity-focused corpora are typically single-label, English-centric, and restricted to broad political categories or explicitly stated group mentions. They rarely offer the fine-grained, multilabel annotations needed to capture the complexity of real-world political speech, and they do not cover non-Latin scripts or languages such as Hebrew. We introduce the first publicly available Hebrew dataset for social identity detection, comprising 5,536 sentences from politicians’ Facebook posts annotated with twelve distinct identities—each grounded in survey-measured salience and expert-defined categories. By providing multilabel annotations across a rich set of ideologically, religiously, and socio-economically relevant identities, our resource enables more nuanced analyses of how political actors deploy identity language than has previously been possible in any non-English setting.

\section{Annotating Social Identities}

\subsection{Panel Survey}

To select social identities for annotation, we utilized a combination of experts and survey instruments. We utilized a representative 12-wave panel survey of the Jewish population in Israel (N = 1,769), conducted between January 2019 and April 2021 (\citealt{Dvir-Gvirsman02112022}; see also \citealt{10.1093/poq/nfaf021}).\footnote{Similar to other panel surveys in which local sample vendors were unable to include re-interview samples of Palestinian citizens in sufficient numbers (e.g., \citet{GIDRON2022102512}), the panel survey did not include this population, reflecting a known deficiency in Israel's survey sampling market.}
This time period contains four Israeli elections held in quick succession, providing a unique opportunity to investigate political dynamics in a condensed time frame.
In 12 survey waves, respondents were asked to select up to three identities they identify with, out of a list of 28. These 28 identities were chosen by a panel of experts in Israeli politics as reflecting a broad spectrum of prevalent social identities, spanning ideological, value-based, religious, national, socio-demographic, and economic dimensions (e.g., Conservative, Leftist, Nationalist, LGBTQ+. See appendix \ref{sec:FullIDlist} for a full list).

For inclusion in the textual annotation, we selected the 12 identities that emerged as most salient in the panel survey, each consistently surpassing a 5\% selection threshold in the first five waves. These identities are Rightist, Leftist, Conservative, Liberal, Socially-oriented, Capitalist, Zionist, Palestinian, Honest, Security-oriented, Ultra-orthodox, and Democrat.\footnote{The 5\% criterion was jointly applied to identities that respondents identify with and identities they disapproved of in a separate survey item.} For more information on the survey methodology, see appendix \ref{sec:survey_methodology}.

%\footnote{Two additional identities (Middle-class and Upper-class) emerged from respondents’ answers but were excluded from the analysis due to empirical limitations: despite repeated refinements, the algorithm consistently failed to detect references to these identities in political discourse.} 

\subsection{Annotation Scheme}

Based on this selection, we developed an annotation scheme for the expression of social identities in text. Identities were annotated only when referenced in a positive manner; negative or oppositional mentions were excluded. Each sentence was annotated using a multilabel scheme across the twelve selected identities.

%Each identity was defined in positive terms, and was only annotated when a speaker referenced an identity in a supportive manner. Negative or critical references (e.g., opposing liberalism or criticizing the ultra-orthodox community) were thus not annotated. Each sentence was annotated in a multilabel manner with respect to the twelve included identities.

%Below is a concise summary of the conceptual definition of each identity used in the annotation alongside relevant abridged examples (full definitions are available in the appendix). 
Below is a summary of the identity definitions used in the annotation scheme, accompanied by examples from the dataset translated from Hebrew (full definitions appear in appendix \ref{sec:CODEBOOK}). Some examples were abridged for brevity.

\paragraph{Liberal:} Advocacy for civil rights, pluralism, separation of religion and state, freedom of religion, protection of minority rights, and support for the judicial system.
\pex 
\textit{\textit{Protecting personal freedom and the right of every individual to be who they are and live as they choose.}}
\xe

\paragraph{Conservative:} Endorsement of opposition to change, support for the integration of religion and state, promotion of anti-liberal values, and advocating for the reduction of judicial authority. 
\pex
\textit{The appointment of conservative and nationalist judges is the most significant factor in changing reality.}
\xe

\paragraph{Democrat:} Emphasis on democratic values and procedures, such as fair elections, the rule of law, and institutional checks and balances. Includes references to the defense of democracy as a political principle.
\pex
\textit{The existence of a democratic, state-based regime in the country; the guarantee of the supremacy of the law.}
\xe

\paragraph{Leftist:} Support for left-wing parties or policies, including dovish security positions, opposition to settlement construction, and criticism of right-wing actors or policies, when tied to a clear ideological stance. \footnote{Note that in Israel, both Leftist and Rightist identities primarily reflect positions on resolving the Israeli-Palestinian conflict, rather than economic issues.}
\pex
\textit{[…] we think that the evacuation of the territories occupied in the Six-Day War is a national necessity of the utmost urgency.}
\xe

\paragraph{Rightist:} Support for right-wing parties or policies, including hawkish security positions, Greater Israel ideology, and criticism of left-wing actors or policies, when tied to a clear ideological stance.
\pex
\textit{[…] we have a true and rare opportunity to form a [fully right-wing] government, so that we can […] pursue an unapologetic policy regarding Israeli settlements in Judea, Samaria, and the Jordan Valley.}
\xe

\paragraph{Capitalist:} Support for free markets, deregulation, private enterprise, reduced government involvement in the economy, and growth-oriented economic policies that avoid redistributive or welfare-based framing.
\pex
\textit{The state should avoid regulatory intervention in the business sector as much as possible}
\xe

\paragraph{Socially-oriented:} Support for social justice and welfare-oriented policies, including references to poverty, job security, government-funded education, healthcare, housing, and accessibility.
\pex
\textit{[…] masses of unemployed, sick, or needy individuals do not place their trust in the free market […], but in the state - in its institutions, its selected officials, and its public servants […].}
\xe

\paragraph{Zionist:} Affirmation of Zionist symbols and values, such as Jewish immigration to Israel, national pride, unity, and collective sacrifice (e.g., references to Memorial Day or the national anthem).%\footnote{This definition draws on the Declaration of Independence, which defines Israel as the homeland of all Jews while committing to full equality for all citizens and minority groups.}
\pex
\textit{It is a great source of pride to see our kindergarten children […] waving the flag with excitement and singing Independence Day songs.}
\xe

\paragraph{Security-oriented:} Focus on national defense, military strength, borders, and threats to the internal or external security of the state. 
\pex
\textit{Only targeting terrorist organizations leaders will create deterrence, protect the security of the residence of the Gaza envelope, and strengthen their resilience.}
\xe

\paragraph{Honest:} References related to corruption and investigations involving public officials, honesty, and ethical conduct.  
\pex
\textit{In two months, we will lead Israel off the path of corruption and onto a new path.}
\xe

\paragraph{Palestinians and Arab Citizens of Israel:} Statements regarding policy issues, worldviews, and ideologies related to Palestinians and Arab-Israelis.
\pex
\textit{We will continue to fight for Arab local authorities and against budgetary discrimination of the Arab society.}
\xe

\paragraph{Ultra-orthodox:} References to the Jewish ultra-Orthodox lifestyle, including the education system, gender segregation, and exemption from military service, as well as mentions of ultra-Orthodox political parties and leadership.

\pex
\textit{[…] it is gratifying to see how the Shas movement […] succeeds in uniting communities through faith in God, the legacy of Rabbi Ovadia Yosef […], and shared values.}
\xe

\subsection{Co-occurring Identities}

Our annotation scheme is multi-label, allowing a single sentence to express several social identities simultaneously. Each annotated sentence is evaluated with respect to all included identities. For example, the following sentence expresses \textit{Democrat}, \textit{Leftist}, and \textit{Honest} social identities:

\pex
\textit{Instead, the current election is about the political survival of a man suspected of deep corruption that endangers Israeli democracy, press freedom, the rule of law, and fairness.}
\xe

Similarly, this sentence expresses both a \textit{Rightist} and a \textit{Liberal} identity:

\pex
\textit{I believe that the nationalist right’s role is to develop settlements everywhere in the country -- without favoritism for one sector or another.}
\xe

\section{Dataset}

To sample sentences for annotation, we compiled a corpus of Facebook posts by Israeli members of parliament, political parties, and viable party candidates, posted during the same timeframe as the survey (Dec. 2018 to Apr. 2021). This corpus comprises 64K posts containing 375K sentences. Our annotated dataset contains 5,536 Hebrew-language sentences sampled from 4.8K unique Facebook posts. Each sentence is annotated with 12 binary annotations according to whether it expresses a given identity. The dataset is split into a training set (70\%), validation set (15\%) and test set (15\%).

\subsection{Inter-coder Reliability}
The sentences were annotated by two of the authors. We evaluate inter-coder reliability on a sample of 304 sentences. The results, given in Table \ref{tab:iaa_scores}, indicate a mean Cohen's $\kappa$ of 0.77. Disagreements were resolved via consultation with all authors.

\begin{table}[ht]
  \centering
  \scriptsize
  \begin{tabularx}{\columnwidth}{%
      >{\raggedright\arraybackslash}l
      *{2}{>{\centering\arraybackslash}X}
    }
    \toprule
    \textbf{Category}
      & \textbf{Percent Agreement}
      & \textbf{Cohen’s $\kappa$} \\
    \midrule
    Conservative          & 96.7\% & 0.705 \\
    Rightist              & 94.4\% & 0.788 \\
    Democrat              & 91.4\% & 0.703 \\
    Honest                & 96.7\% & 0.782 \\
    Capitalist            & 98.4\% & 0.792 \\
    Ultra-Orthodox        & 98.7\% & 0.708 \\
    Socially-oriented     & 96.1\% & 0.841 \\
    Liberal               & 96.1\% & 0.841 \\
    Leftist               & 94.1\% & 0.761 \\
    Security-oriented     & 96.4\% & 0.736 \\
    Palestinian           & 98.4\% & 0.792 \\
    Zionist               & 96.1\% & 0.778 \\
    \midrule
    \textbf{Mean}         & 96.1\% & 0.769 \\
    \bottomrule
  \end{tabularx}
    \caption{Inter-Annotator Agreement (IAA) scores}
  \label{tab:iaa_scores}
\end{table}

\subsection{Descriptive statistics}
The number of positive labels per identity ranges between 129 (Ultra-Orthodox) and 703 (Rightist), with a mean of 413 (Table~\ref{tab:positive_counts}). 37.5\% of sentences express no identity, 41.1\% express one identity and 21.4\% express two or more identities (Table \ref{tab:positives_distribution}).

\begin{table}[ht]
  \centering
  \small

  \begin{tabular}{@{}l r@{}}
    \toprule
    \textbf{Category}   & \textbf{Count} \\
    \midrule
    Rightist            & 703 \\
    Liberal             & 629 \\
    Socially-oriented   & 567 \\
    Democrat            & 562 \\
    Leftist             & 546 \\
    Zionist             & 368 \\
    Honest              & 357 \\
    Security-oriented   & 346 \\
    Conservative        & 297 \\
    Capitalist          & 230 \\
    Palestinian         & 225 \\
    Ultra-Orthodox      & 129 \\
    \midrule
    \textbf{Mean}       & 413.25 \\
    \bottomrule
  \end{tabular}
    \caption{Number of Positive Instances}
  \label{tab:positive_counts}
\end{table}

\begin{table}[ht]
  \centering

  \begin{tabular}{crr}
    \toprule
    \textbf{\# Positives} & \textbf{Row Count} & \textbf{Percent} \\
    \midrule
    0 & 2,078 & 37.54\% \\
    1 & 2,275 & 41.09\% \\
    2 &   919 & 16.6\%  \\
    3 &   218 & 3.94\%  \\
    4 &    39 & 0.7\%   \\
    5 &     6 & 0.11\%  \\
    6 &     1 & 0.02\%  \\
    \bottomrule
  \end{tabular}
    \caption{Number of Positive Labels per Sentence}
  \label{tab:positives_distribution}
\end{table}

\section{Modeling and Experiments}

We model the task of identifying social identities from text using four types of training setups:

First, we establish \textbf{Baseline models} by training separate Logistic Regression and LinearSVC classifiers for each label using TF-IDF features. Then, \textbf{Multilabel encoder-models} are fine-tuned to jointly learn the 12 identity labels.
\textbf{Single-label encoder models} are fine-tuned separately for each label. For this training, we utilize the best performing base model in the multilabel encoder setup.
Finally, \textbf{Decoder LLMs} are fine-tuned to generate a comma-separated list of all applicable Hebrew language labels for the given sentence.

We examine a set of encoder models and LLM decoders in the 2B--9B parameter range. For encoders, we use the multilingual \textsc{mBERT} \citep{devlin2019bert}, and Hebrew-targeted encoders \textsc{AlephBERT} \citep{seker-etal-2022-alephbert}, \textsc{HeRO} \citep{shalumov2023herorobertalongformerhebrew} and the base and large variants of \textsc{DictaBERT} \citep{shmidman2023dictabert}. All encoder models are trained for up to 10 epochs with three learning rates (1e-5, 3e-5, 5e-5), and three types of loss (default, positive weight and focal loss), choosing the best performing checkpoint on the validation set. 

For decoder LLMs, we use \textsc{Gemma 2} in the 2B and 9B variants \citep{gemma2024}, \textsc{Qwen3-8B} \citep{qwen3}, and the Hebrew-targeted \textsc{DictaLM2.0} \citep{shmidman2024adaptingllmshebrewunveiling}. All decoders are fine-tuned with QLORA \citep{dettmers2023qlora} for up to 5 epochs with two learning rates (1e-5, 1e-4).

\begin{table*}[ht]
  \centering
  \small
  \renewcommand{\arraystretch}{1.1}

  \resizebox{\linewidth}{!}{%
    \begin{tabular}{@{}l | c c c | c c c c c c c c c c c c@{}}
      \toprule
  %    Model            & P     & R     & F$_1$ 
  %                     & Bit   & Cap   & Con   & Dem   & Har   & Hev   & Lib   
  %                     & Pal   & Smo   & Yem   & YD    & Zio   
        & \multicolumn{3}{c}{\textbf{Macro-averaged metrics}}
      & \multicolumn{12}{c}{\textbf{Per-label F$_1$ scores}} \\
      \cmidrule(lr){2-4}\cmidrule(lr){5-16}
        Model      & P    & R    & F$_1$ 
                 & \makecell[c]{Security-\\oriented}
                 & \makecell[c]{Capitalist}
                 & \makecell[c]{Conserva\\-tive}
                 & \makecell[c]{Democrat}
                 & \makecell[c]{Ultra-\\Orthodox}
                 & \makecell[c]{Socially-\\oriented}
                 & \makecell[c]{Liberal}
                 & \makecell[c]{Palestinian}
                 & \makecell[c]{Leftist}
                 & \makecell[c]{Rightist}
                 & \makecell[c]{Honest}
                 & \makecell[c]{Zionist} \\
  \\
      \midrule
      \multicolumn{16}{@{}l}{\textbf{Decoder-only}}                                                            \\
      \textsc{DictaLM2.0}       & \textbf{0.740} & \textbf{0.751} & \textbf{0.743} 
                       & \textbf{0.705} & \textbf{0.805} & 0.675 & 0.754 & 0.653 & \textbf{0.723} & 0.724 
                       & \textbf{0.852} & \textbf{0.750} & \textbf{0.765} & \textbf{0.816} & \textbf{0.700} \\
      \textsc{Gemma-2-9B}         & 0.717 & 0.698 & 0.705 
                       & 0.645 & 0.759 & \textbf{0.684} & 0.753 & 0.591 & 0.662 & \textbf{0.749} 
                       & 0.758 & 0.671 & 0.737 & 0.804 & 0.650 \\
      \textsc{Gemma-2-2B}         & 0.620 & 0.631 & 0.624 
                       & 0.560 & 0.723 & 0.575 & \textbf{0.757} & 0.385 & 0.634 & 0.673 
                       & 0.716 & 0.609 & 0.615 & 0.680 & 0.557 \\
      \textsc{Qwen-8B}          & 0.665 & 0.463 & 0.542 
                       & 0.605 & 0.508 & 0.308 & 0.700 & 0.410 & 0.541 & 0.568 
                       & 0.625 & 0.516 & 0.515 & 0.650 & 0.554 \\
      \midrule
      \multicolumn{16}{@{}l}{\textbf{Multilabel encoders}}                                                      \\
      \textsc{DictaBERT}\textsubscript{Large}      & 0.677 & 0.680 & 0.678 
                       & 0.660 & 0.667 & 0.561 & 0.750 & \textbf{0.667} & 0.688 & 0.689 
                       & 0.831 & 0.588 & 0.670 & 0.716 & 0.645 \\
      \textsc{DictaBERT}$_{\text{Base}}$      & 0.629 & 0.710 & 0.664 
                       & 0.667 & 0.667 & 0.511 & 0.753 & 0.533 & 0.659 & 0.697 
                       & 0.806 & 0.663 & 0.657 & 0.750 & 0.602 \\
      \textsc{AlephBERT}        & 0.628 & 0.692 & 0.657 
                       & 0.673 & 0.675 & 0.526 & 0.738 & 0.604 & 0.627 & 0.664 
                       & 0.783 & 0.620 & 0.641 & 0.718 & 0.618 \\
      \textsc{HeRo}             & 0.608 & 0.693 & 0.647 
                       & 0.667 & 0.575 & 0.587 & 0.730 & 0.528 & 0.648 & 0.642 
                       & 0.776 & 0.659 & 0.638 & 0.720 & 0.589 \\
      \textsc{mBERT}            & 0.573 & 0.553 & 0.552 
                       & 0.483 & 0.467 & 0.527 & 0.708 & 0.356 & 0.568 & 0.635 
                       & 0.632 & 0.531 & 0.502 & 0.653 & 0.566 \\
      \midrule
      \multicolumn{16}{@{}l}{\textbf{Single-label encoders}}                                                    \\
      \textsc{DictaBERT}\textsubscript{Large}  & 0.643    & 0.689    & 0.659 
                       & 0.672 & 0.659 & 0.529 & 0.731 & 0.508 & 0.708 & 0.694 
                       & 0.783 & 0.615 & 0.636 & 0.721 & 0.652 \\
      \midrule
      \multicolumn{16}{@{}l}{\textbf{Baselines}}                                                                 \\
      LinearSVC       & 0.398    & 0.339    & 0.361
                      & 0.400 & 0.170 & 0.300 & 0.490 & 0.230 & 0.430 & 0.330
                      & 0.520 & 0.390 & 0.390 & 0.350 & 0.350 \\
      LogisticRegression & 0.469    & 0.268    & 0.334
                      & 0.350 & 0.000 & 0.280 & 0.500 & 0.280 & 0.410 & 0.340
                      & 0.470 & 0.380 & 0.350 & 0.360 & 0.290 \\
      \bottomrule
    \end{tabular}

  }%
        \caption{Macro-averaged precision (P), recall (R), F$_1$, and per-label F$_1$ for all models.}
  \label{tab:perf}
\end{table*}

Table~\ref{tab:perf} reports test set results for trained models. We observe that the most performant are decoder models, achieving the highest per-label F1 scores on all but one label. \textsc{DictaLM2.0}, specifically, achieves the best macro P/R/F1 scores, and best per-label scores on 8/12 identities. The performance of this model, continuously pre-trained from Mistral 7B on 100B Hebrew tokens, underscores the dependence identity detection on language specific cultural and political context.

We also note that two of the multilabel encoders, \textsc{DictaBERT}\textsubscript{Large} and \textsc{DictaBERT}$_{\text{Base}}$, perform better on average than the single-label encoders fine-tuned for each identity separately using \textsc{DictaBERT}\textsubscript{Large} as a base model. \textsc{DictaBERT}\textsubscript{Large} achieved an $F_1$ score of 0.678 in a multi-label setting, compared to a mean $F_1$ of 0.659 when combining the single-label models. This indicates the benefit of joint learning and the interrelated nature of identities in this task.

\subsection{Generalization to Parliamentary Speech}

We examine whether the resulting classifier generalizes well to the parliamentary speech in the Knesset by utilizing the IsraParlTweet dataset \citep{mor-lan-etal-2024-israparltweet}. We subset speeches from the equivalent time period of the Facebook data, classify all sentences with the best performing \textsc{DictaLM2.0} model, and sample 500 of the sentences for human annotation. We perform a precision-oriented test by oversampling positive predictions for each identity. The results show a macro $F_1$ score of 0.72, on par with the Facebook test set, indicating the model's ability to generalize to Knesset data (for full results, see Table \ref{tab:knesset_generalization}).

\subsection{External Validity}

We examine the external validity of the classifier by correlating social identity discourse  with external party policy rankings from the CHES-Israel expert survey \citep{doi:10.1177/13540688231218917}, which assigns parties scores on several policy domains. For each of the 16 political parties covered by both the expert survey and our Facebook corpus in year 2021, we calculate identity discourse scores, defined as the share of sentences posted by the party's politicians which are classified as expressing said identity.

We match closely related policy issues from the expert survey to each social identity. For example, policy issue \textit{Economic Stance} is linked with both \textit{Capitalist} and \textit{Socially-oriented} social identities. For better comparability, in some cases we additionally construct social identity dimensions by subtracting party means for two polar identities (e.g. $\text{Rightist} - \text{Leftist}$, $\text{Capitalist} - \text{Socially-Oriented}$). We then compute correlations between the CHES party policy scores and the predicted social identity discourse scores, for the 16 political parties.

Our analysis reveals a strong alignment with the expert-coded CHES data. Out of 21 correlations, 16 are statistically significant at $p \le 0.1$ and 13 are significant at $p \le 0.05$. Of those 13, all fall within a strong effect size range, with absolute correlation values from $|r| = 0.71$ to $0.94$. For full results, see Table~\ref{tab:ches_correlation} in the appendix.

\section{Results}

We utilize the three identity data sources with overlapping time frames: our corpus of Israeli politician Facebook posts; the subset of parliamentary speeches from \citep{mor-lan-etal-2024-israparltweet} in the same time period; and the panel survey capturing the public's identity choices. The first two sources are classified using the best performing \textsc{DictaLM2.0} model. Utilizing these three data sources allows for a unique exploration of the dynamics of social identities on different platforms of elite discourse (Facebook and Knesset speeches) and the public (panel survey). We examine the following aspects: the popularity of identities; temporal trends; the correlation and bundling together of identities; and gender differences in identities.

\paragraph{Popularity.}

\begin{figure*}
    \centering
    \includegraphics[width=0.8\linewidth]{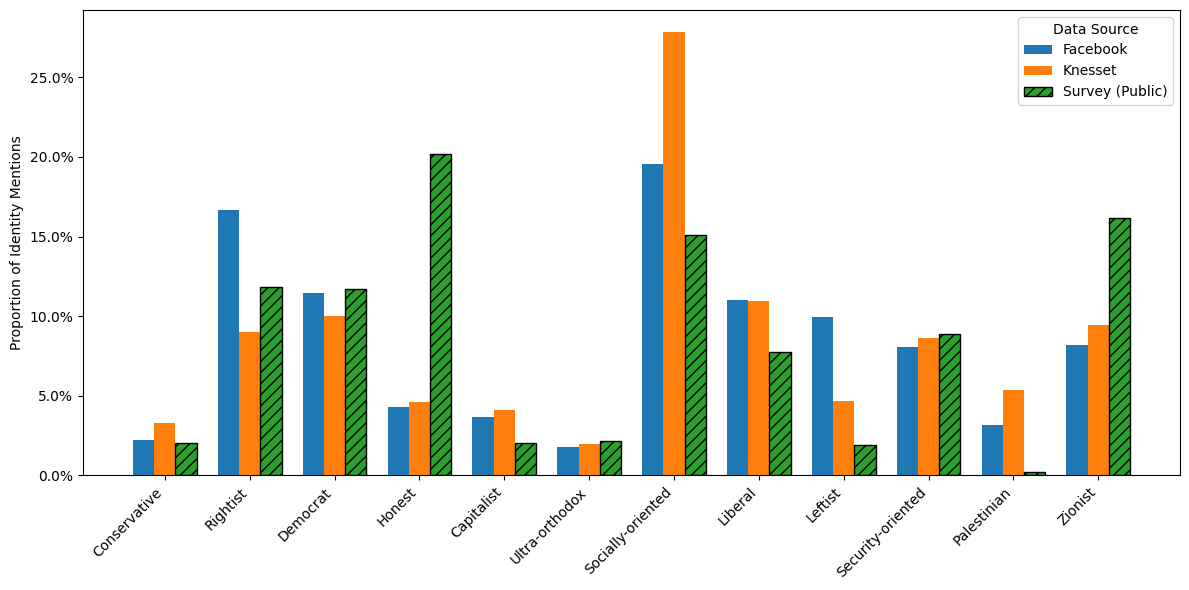}
    \caption{Normalized share of identities}
    \label{fig:top_identities}
\end{figure*}

In Figure~\ref{fig:top_identities}, we compare the normalized share of each identity. The identities \textit{Socially-oriented}, \textit{Rightist} and \textit{Democrat} are generally popular across the data sources. Interestingly, the \textit{Leftist} identity is relatively less popular. While many identities receive similar proportions among the three sources, several exceptions stand out. Identities \textit{Honest} and \textit{Zionist} are significantly more popular among the public, whereas \textit{Socially-oriented} is significantly more popular in parliament.

Examining correlations between the ranks of identities in each data source, we see that while the popularity ordering of identities is strongly correlated between the Facebook and Knesset data ($\rho$=0.87), survey identity rankings are moderately correlated with Facebook ($\rho$=0.48) and Knesset ($\rho$=0.46) data. For all ranks, see Figure ~\ref{fig:identity_ranking}.

\paragraph{Temporal trends.}

\begin{figure*}
    \centering
    \includegraphics[width=0.8\linewidth]{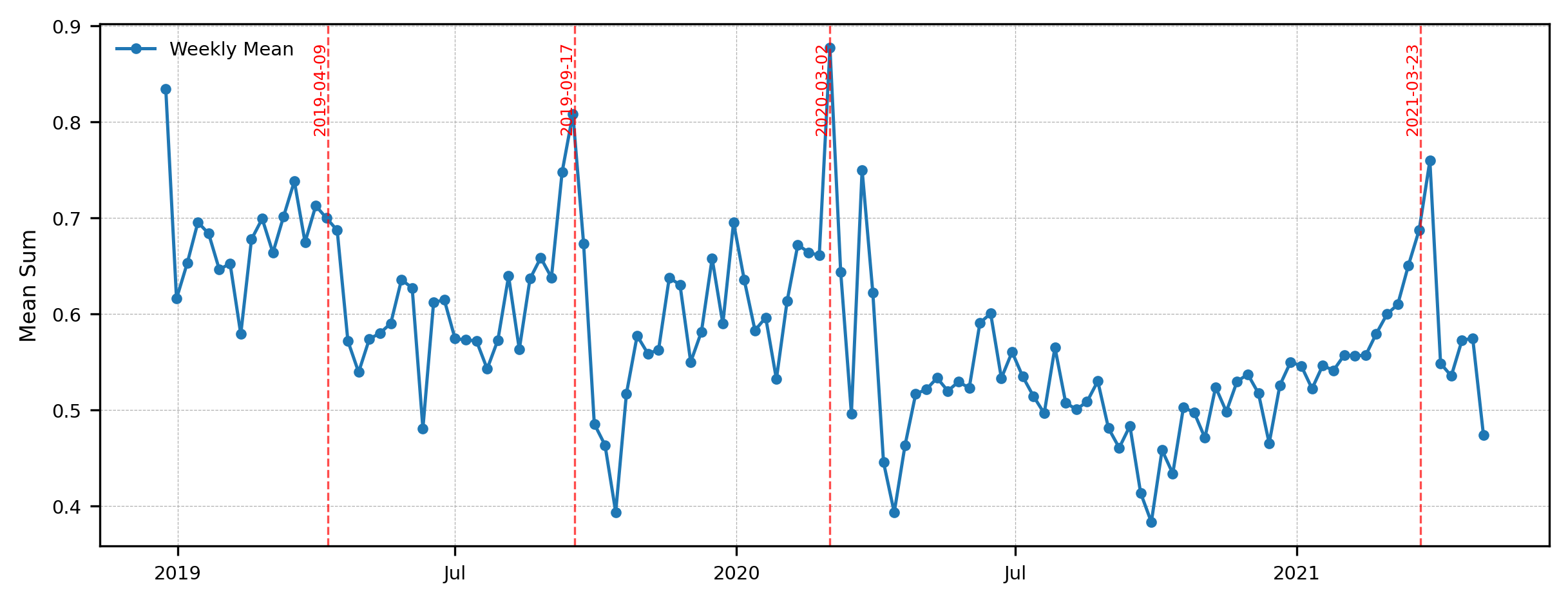}
    \caption{Temporal Trend}
    \label{fig:temporal_trend}
\end{figure*}

We examine whether identity-related discourse tends to increase before elections. Facebook posts are the most suitable data source for this analysis, as the survey panel includes a limited number of waves and Knesset meetings are in recess near election periods. We plotted the average number of identity mentions per sentence per week in the Facebook data (see Figure \ref{fig:temporal_trend}). The results show considerable fluctuation, ranging from 0.38 to 0.87 identity mentions per sentence. Notably, identity-related discourse peaks around three of the four election dates, highlighting the salience of identities during electoral competition.

\begin{figure*}
    \centering
    \includegraphics[width=0.8\linewidth]{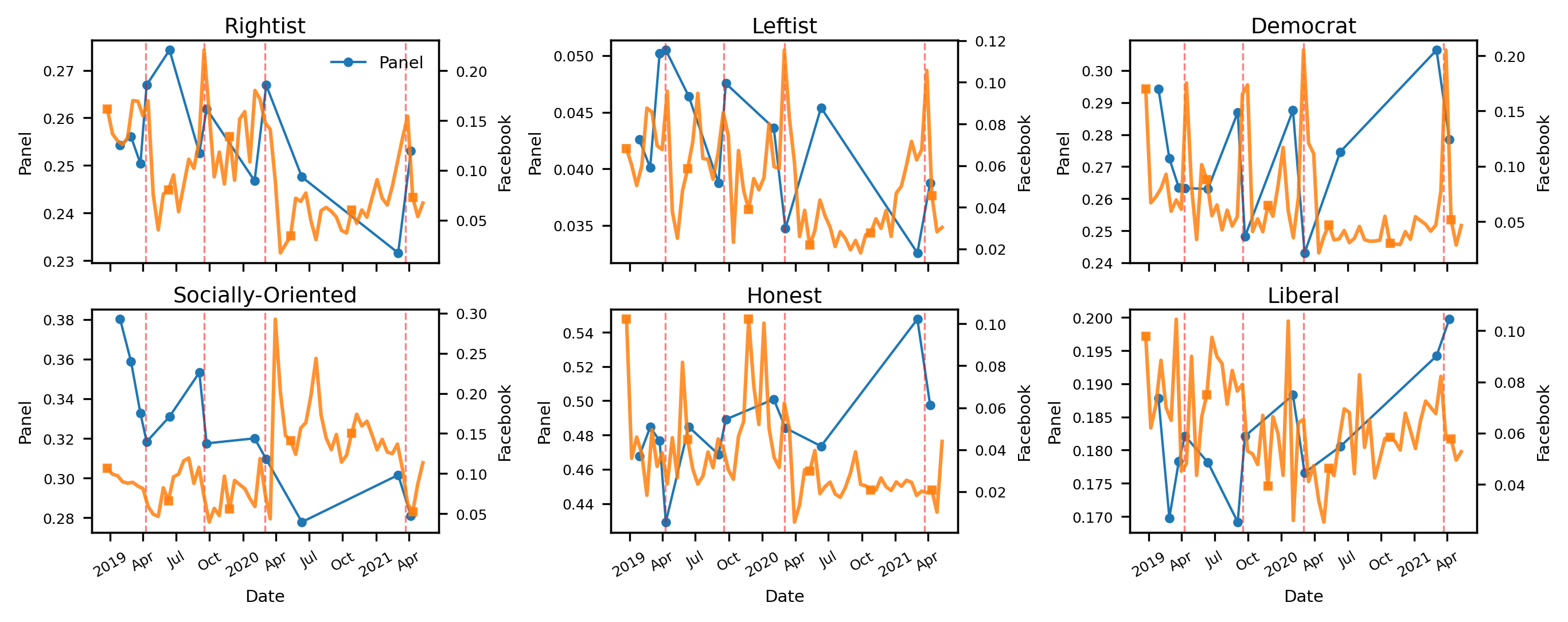}
    \caption{Per Identity Trends - Facebook and Survey}
    \label{fig:per_identity_six}
\end{figure*}

Figure~\ref{fig:per_identity_six} plots the share of six identities over time for both the Facebook and survey samples (see Figure \ref{fig:identity_timetrends} for all 12). We see that the identities that peak during or near election dates are \textit{Rightist}, \textit{Leftist} and \textit{Democrat}. However, as previously mentioned, \textit{Leftist} holds a very small share in both Facebook and the survey, and thus drives a smaller part of the overall temporal change. A notable gap between elites and the public emerges for three identities. While the public decreasingly identifies with the \textit{Socially-oriented} identity, it becomes significantly more prominent among politicians after the 3rd elections held in March 2020. On the other hand, the identities \textit{Honest} and \textit{Democrat} become more popular among the public after the 3rd election, but less popular in elite discourse. These findings demonstrate the potential of \textsc{HebID} to provide valuable insights into how identity-driven agendas shift in the lead-up to elections.

\begin{figure*}[!htb]
    \centering
    \includegraphics[width=1\linewidth]{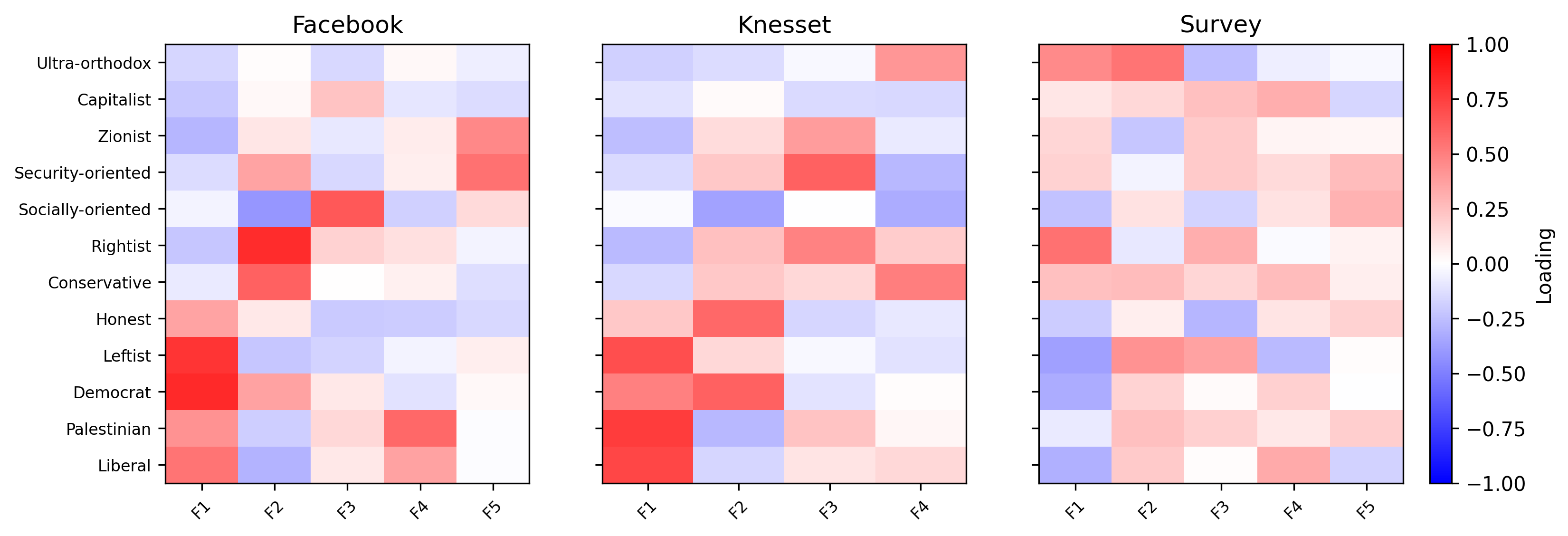}
    \caption{Factor analysis}
    \label{fig:factor-analysis}
\end{figure*}

\paragraph{Identity bundles.}
To examine which identities tend to co-occur, we perform a factor analysis on the mean levels of each identity per speaker/respondent (Figure \ref{fig:factor-analysis}). In the three sources, we see a primary dimension dividing identities into two broad groups, a left-wing group (\textit{Leftist}, \textit{Democrat}, \textit{Honest}, \textit{Liberal}, \textit{Palestinian}) and a right-wing group (\textit{Rightist}, \textit{Conservative}, \textit{Zionist}, \textit{Security-oriented}, \textit{Capitalist}, \textit{Ultra-orthodox}). Other factors in the Facebook and Knesset data appear to reflect political sub-dimensions of identity pairs that are more tightly coupled, such as \textit{Conservative} and \textit{Rightist}, \textit{Zionist} and \textit{Security-oriented}, \textit{Liberal} and \textit{Palestinian}, and \textit{Honest} and \textit{Democrat}. %Additional factors in the survey data appear to have less structure.
%shaul alternative: Factors in the survey data appear less structured compared to political elites.
In the survey data, factors beyond the first dimension appear less structured compared to those that arise in the Facebook and Knesset datasets.

Accordingly, examination of the correlations between identities (after aggregating into speakers/respondents) shows that the survey exhibits a mean absolute correlation coefficient of 0.159, weaker than the Facebook sample (0.235) and Knesset data (0.215). For full correlation matrices, see appendix \ref{sec:additional_results}.

\paragraph{Gender differences.}
Do men and women differ in terms of social identities? We first examine gender differences in identity discourse by calculating gender differences in the total number of identities expressed. For each speaker/respondent, we calculate the mean number of identities expressed in a sentence or survey response, and then aggregate by gender. We see that in all data sources, women express more identities than men. However, the gap is largest in the Knesset data (0.07) and is not statistically significant in the survey data.

We then examine the gender difference per identity by subtracting the share of each identity among women from that of men, in all data sources. We see that some identities, such as \textit{Rightist}, \textit{Security-oriented}, \textit{Capitalist}, and \textit{Ultra-orthodox} lean towards men, whereas \textit{Socially-oriented} significantly leans towards women across all platforms. While the gender gaps on Facebook and the Knesset are generally in the same direction, several of the survey gender gaps arise in different directions. Thus, \textit{Honest} in the survey leans significantly towards women, whereas in the Facebook and Knesset data it leans slightly towards men.

\begin{figure*}[!h]
    \centering
    \includegraphics[width=0.8\linewidth]{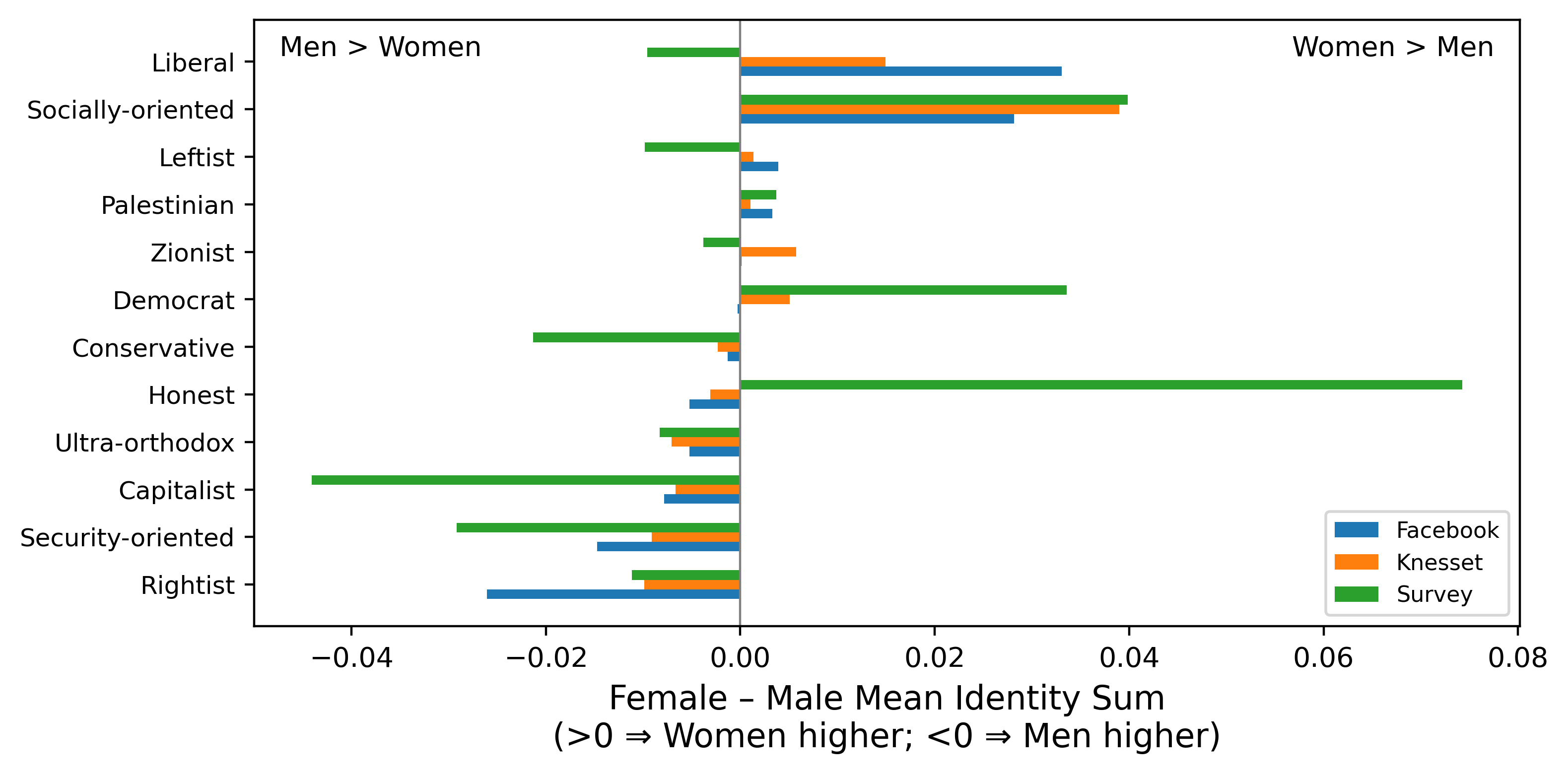}
    \caption{Gender Difference per Identity}
    \label{fig:enter-label1}
\end{figure*}

\begin{figure}[!h]
    \centering
    \includegraphics[width=1\linewidth]{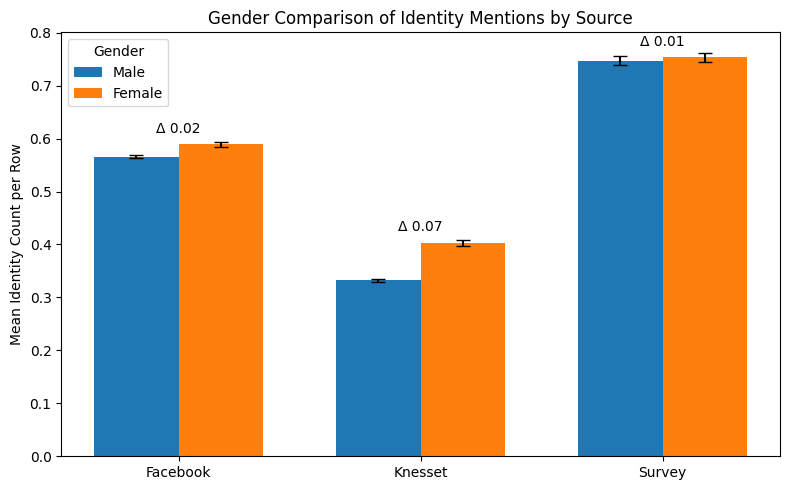}
    \caption{Gender Difference in Identity Mentions}
    \label{fig:enter-label}
\end{figure}

\section{Conclusion}
In this work, we have introduced \textsc{HebID}, the first publicly available multilabel Hebrew corpus for fine-grained social identity detection in political text, grounded in large-scale survey evidence and expert consultation. Our dataset of 5,536 sentences from Israeli politicians’ Facebook posts is annotated for twelve empirically salient identities, ensuring both linguistic and sociological validity. We benchmarked a range of models—multilabel and single-label encoders as well as 2B–9B-parameter decoder LLMs—and demonstrated that Hebrew-tuned decoder models (e.g., \textsc{DictaLM2.0}) achieve the highest macro-F1 (0.743), significantly outperforming encoder-only baselines. We further validated cross-genre robustness by applying our best model to 500 Knesset speech excerpts, matching social-media performance (macro-F1 = 0.72). 

By linking three complementary sources — Facebook posts, parliamentary speeches, and a national survey — we revealed systematic differences in identity prevalence, temporal surges around elections, bundles of co-occurring identities, and gendered patterns of identity expression. These findings showcase the potential of our framework to uncover nuanced political dynamics and address substantive questions in the social sciences.

The \textsc{HebID} annotation scheme and corpus provides a comprehensive foundation for future research on identity discourse in non-English political contexts and paves the way for comparative studies across political systems. Specifically, this work provides \textit{a new tool for social scientists} studying political communication, \textit{a new benchmark for NLP researchers} working on Hebrew and multilabel classification, and \textit{a methodological template} for creating similar resources cross-culturally.

\clearpage
\section*{Limitations}
While \textsc{HebID} represents a significant step forward in Hebrew political text analysis, several limitations should be acknowledged:
\begin{itemize}
  \item \textbf{Temporal and platform scope:} 
  %Our annotations cover Facebook posts from December 2018 to April 2021 and Knesset speeches in the same period. Identity discourse may have evolved since our data collection, and other platforms (e.g., Twitter, news media) are not represented.
  Identity discourse is dynamic and may have evolved beyond the period captured in our data. Additionally, other platforms—such as Twitter and news media—are not represented, leaving out potentially important dimensions of identity expression and change over time.

  \item \textbf{Survey population:} The panel survey sampled only Jewish citizens of Israel. Identities and salience patterns may differ among non-Jewish citizens, including Palestinians and non-Jewish immigrant communities.  
  \item \textbf{Annotation granularity:} Although multilabel, our scheme relies on twelve categories selected via a 5\% survey threshold. Less frequent but potentially important identities were excluded, and negative or critical references to an identity are not captured.
  \item \textbf{Model biases:} Our classifiers inherit biases present in both the training data and the pretrained language models (e.g., under- or over-representation of certain groups). Performance may degrade on dialectal text, informal registers, or in hostile political discourse.
  \item \textbf{Cross-genre validation:} The Knesset evaluation, while indicative of robustness, is based on a limited sample of 500 human-annotated sentences drawn from Knesset Plenum protocols. A broader evaluation across other legislative bodies or timeframes is needed to fully assess generalization.
  \item \textbf{Methodological considerations:} Survey responses and text analysis offer complementary but distinct measurement approaches. Survey data provide insights into self-reported identity salience, while text analysis reveals patterns of expressed identity discourse. These methodological differences suggest caution when comparing panel survey results with computational text analysis findings.
\end{itemize}
Future work should extend \textsc{HebID} to additional platforms and populations, refine the annotation taxonomy to include emerging identities, and explore methods to mitigate model bias in identity detection tasks.

\section*{Acknowledgments}

This work was supported by the Israel Science Foundation (Grant no. 2501/22). We thank Vered Porzycki for assistance in conceptualizing the definitions of the annotated identities; Vered Porzycki, Gal Ron, Avishai Green, Dror Markus and Noa Amir for invaluable assistance in a preliminary pilot study; and Effi Levi for his very useful and insightful advice.

\bibliography{custom}

\appendix

\section{Full list of social identities in Survey}
\label{sec:FullIDlist}

\begin{enumerate}
    \item Ashkenazi (Jewish of European descent)
    \item Capitalist
    \item Conservative
    \item Democrat
    \item A man of faith
    \item Honest
    \item Humanist
    \item Israeli
    \item Jewish
    \item Leftist
    \item LGBTQ+
    \item Liberal
    \item Lower class
    \item Man
    \item Middle class
    \item Mizrahi (Jewish of Middle Eastern/North African descent)
    \item Nationalist
    \item Palestinian
    \item Religious
    \item Rightist
    \item Secular
    \item Security-oriented
    \item Socialist
    \item Socially-oriented
    \item Ultra-Orthodox (Haredi)
    \item Upper class
    \item Woman
    \item Zionist
\end{enumerate}

\section{Survey Methodology}
\label{sec:survey_methodology}

The survey we utilize is an online panel survey conducted between January 2019 and April 2021, using a representative sample of the adult Jewish population in Israel. Participants were recruited through iPanel, the largest online survey firm in Israel, and received gift cards as compensation. To ensure representativeness, respondents were selected using demographic quotas based on key population parameters: age, gender, education, geographic location, and religiosity. A total of 1,769 individuals completed the first wave in January 2019, and 878 panelists completed the relevant waves for this study.

To measure identity salience, respondents were asked the following question: \textit{"People feel a sense of belonging to some groups in society and a sense of distance from others. Below is a list of groups. Please select up to three groups that best define who you are, groups that make you feel a sense of belonging, identification, and pride, and that you would like to see properly represented in the Knesset."}

The possible selection options are the 28 identities appearing in section \ref{sec:FullIDlist} of the appendix, a list developed by experts in Israeli politics to ensure relevance and comprehensiveness.

In total, the panel included 14 waves and an additional minor wave in which only several question items were repeated. We utilize 12 of the waves in which the identity salience question appears (11 major waves and the minor wave).

\section{Identity Definitions: Codebook}
\label{sec:CODEBOOK}

\noindent Dear Coder,\\
Below are twelve identity definitions for classification. Please follow the following guidelines:

\begin{enumerate}
    \item Classify Based on Definitions. Assign identity categories only to statements that align with the definitions provided.
    \item Multiple Classifications. If a statement may fit more than one identity category. Please assign it to all relevant identity categories accordingly.
    \item Positive Associations Only. Please note that the identity categories reflect a positive association. Do not classify statements that oppose an identity (e.g., criticism of capitalism, liberalism, Ultraorthodox) under that category.
    \item The Speaker Identity Is Irrelevant. Classification is based on the content of the statement, not who says it (e.g., a non-Palestinian can make a statement classified under ``Palestinians and Arab Citizens of Israel'' if they present a Palestinian perspective).
    \item Not all statements will be classified. Some statements may not fit any of the twelve identity categories mentioned in the codebook. In such cases, do not force a classification.
\end{enumerate}

\noindent\textbf{Identity definitions:}\\

\noindent\textbf{Capitalist.} Support for private enterprises, free markets and economics (including the stock market), wealth accumulation, trade-related topics, deregulation policies, tax reductions, limiting government involvement in the economy or social services. General references to economic growth, market openness, or business encouragement—without mention of state intervention—will be labeled under this identity.

\noindent\textbf{Socially-oriented (\textit{Hevrati}).} Support for social justice and welfare-oriented policies, including reference to poverty in Israel, job security, healthcare, government-funded education, social organizations, aid initiatives (e.g., food donations), social policy (e.g., housing assistance, support for disadvantaged populations), social security benefits, infrastructure investment, prioritizing local business over international ones, and environmental policy.

\noindent\textbf{Conservative.} Endorsement of opposition to change, support for the integration of religion and state, preservation of traditional values, promotion of anti-liberal values, and advocating for the reduction of judicial authority. Note: This identity includes reference that endorse Jewish traditions and opposition to anti-religious coercion.

\noindent\textbf{Liberal.} Advocacy for equality, human and civil rights, pluralism, separation of religion and state, freedom of religion, the promotion of universal values, protection of minority or LGBTQ+ rights, and support for the judicial system.\\
Note that:
\begin{enumerate}
    \item References relating to inequality in the burden of civic duties will be labeled under this identity.
    \item This identity does not include references to the rule of law.
\end{enumerate}

\noindent\textbf{Democrat.} Emphasis on democratic values and procedures, such as fair elections, the rule of law, institutional checks and balances (``rules of the game''), and the defense of democracy as a political principle. Includes explicit references to democracy or portraying Israel as a democratic state.

\noindent\textbf{Honest (\textit{Yeshar Derech}).} References related to corruption and investigations involving public officials, honesty, integrity, and ethical conduct in public service. Includes mentions of improper immunity, criticism of self-serving behavior by public officials, and praise for individuals acting in public interest. Note that neutral factual statements (e.g., ``Netanyahu's trial begins tomorrow in Jerusalem'') and descriptive mentions of corruption without a clear perspective will not be classified under this identity.

\noindent\textbf{Leftist.} Support for left-wing parties or policies, including dovish security policies, willingness to make territorial compromise, opposition to settlement construction, criticism of right-wing actors or policies, when tied to a clear ideological stance.\\
Note that:
\begin{enumerate}
    \item Identification with left-wing parties or groups is labeled as Leftist.
    \item This identity focuses on hawkish-dovish positions, not economic or social issues (which are coded separately, e.g., Liberal or Socially-oriented).
    \item References to ``peace'' alone do not constitute a leftist identity without additional ideological context.
    \item Centrist parties opposing a side are labeled according to the target of opposition (e.g., ``Blue and White opposes Likud'' = Leftist; ``Blue and White opposes Labor'' = Rightist).
    \item Positions regarding state-religion relations are not considered part of the left/right ideological identities (e.g., referring to religious coercion would be considered liberal, not Leftist).
\end{enumerate}

\noindent\textbf{Rightist.} Support for right-wing parties or policies, including hawkish security positions, Greater Israel ideology, criticism of left-wing actors or policies, when tied to a clear ideological stance.\\
Note that:
\begin{enumerate}
    \item Identification with right-wing parties or groups is labeled as Rightist.
    \item This identity focuses on hawkish-dovish positions, not economic or social issues (which are coded separately, e.g., Conservative or Capitalist).
    \item References to Israel as a ``Jewish and democratic state'' do not alone indicate Rightist identity.
    \item Centrist parties opposing a side are labeled according to the target of opposition (e.g., ``Blue and White opposes Labor'' = Rightist; ``Blue and White opposes Likud'' = Leftist;).
    \item Positions regarding state-religion relations are not part of ideological identities (e.g., referring to greater congruence between state and religious laws would be considered Conservative not Rightist).
\end{enumerate}

\noindent\textbf{Palestinians and Arab Citizens of Israel.} Statements regarding policy issues, worldviews, and ideologies that represent or reflect Palestinians and Arab-Israelis, including references to Palestinian and Arab-Israeli culture, statements and interviews by Palestinian and Arab-Israeli public leaders related to Palestinians and Arab-Israelis, as well as policy matters concerning these groups (not just security issues).\\
Note that:
\begin{enumerate}
    \item References that do not reflect a Palestinian perspective (e.g., discussions on settlements or military actions against Palestinian organizations), will not be labeled under this identity.
    \item The speaker does not have to be Palestinian but must present events or impacts from a Palestinian perspective.
    \item Positive references for public figures representing these communities will be included under this identity.
    \item This identity is distinct from the Leftist identity. References can be to both or either, depending on context.
\end{enumerate}

\noindent\textbf{Security-Oriented (\textit{Bitchonist}).} References to national security issues, military strength, security capabilities, threats to internal or external safety, defense agencies (IDF, Shin Bet, Mossad, etc.), borders protection, the state's ability to protect its citizens and security challenges.\\
Note that:
\begin{enumerate}
    \item This identity includes references that relate to the notion of protecting all citizens (e.g., Jewish, Arab-Israeli) from violence or terror.
    \item This identity excludes general or symbolic mentions of soldiers unrelated to defense (e.g., prayers, greetings, daily life).
    \item This identity includes critiques of external actors (e.g., referring to the Palestinian Authority as a terrorists funding organization) when framed in terms of national security.
%    \item ``No peace without security'' statements fall under this identity.
\end{enumerate}

\noindent\textbf{Ultra-orthodox (\textit{Haredi}).} References to Jewish ultra-orthodox (Haredi) lifestyle, including the Haredian education system, gender segregation in the public sphere, exemption from military service for yeshiva students, charitable activities within the Haredi community, the Haredi religious (Torah) world and tradition preservation, Haredian parties (Shas, United Torah Judaism, Agudat Yisrael), and Haredi rabbinic leadership.\\
Note that:
\begin{enumerate}
    \item The mere mentioning of Rabbis is not enough. Statements must carry a positive tone.
    \item General religious expressions (e.g., quoting a verse, mentioning Jewish holidays) will not be labeled under this identity unless clearly framed within Haredian context or authority.
    \item References to the integration of Haredim into broader Israeli society, or criticism of the Haredi community, will not be labeled under this identity.
\end{enumerate}

\noindent\textbf{Zionist.} Affirmation of Zionist symbols and values, such as Jewish immigration to Israel (Aliyah), connections between Israel and the Jewish diaspora, national pride, IDF enlistment, national unity, symbolic expressions such as the positive references to the Israeli flag, national anthem and collective sacrifice (e.g., references to Memorial Day, families of fallen soldiers). Also includes references relating to Israel's struggle against antisemitism and BDS.\\
Note that:
\begin{enumerate}
    \item This definition draws on the Declaration of Independence, which defines Israel as the homeland of the Jewish people.
    \item This identity includes expressions of dedication to Israel's future, strength and prosperity (``to serve the country,'' ``for the future of the country'').
  %  \item This identity does not include general mentions of antisemitism without connection to Israel, references to soldiers unrelated to enlistment or national service.
\end{enumerate}

\section{Fine-tuning setup}

All model fine-tuning is performed on an 80GB A100 Nvidia GPU, using huggingface transformers.

For decoder fine-tuning, the separator "\#\#\# Answer:" is used to separate the input sentences from the output. The labels of the input and separator are loss-masked.

All experiments utilize AdamW optimizer with a linear scheduler. Default values of hyperparameters are used everywhere except for learning rate (for encoders and for decoder LLMs) and loss type (for encoder models).

Decoder fine-tuning uses QLORA with 4bit quantization. LORA settings are rank of 256, and alpha value of 512. LORA layers are attached to all linear levels in the decoder models.

Each hyper-parameter configuration was trained once.

\section{Generalization to Parliamentary Speech }

Table \ref{tab:knesset_generalization} shows the full results of the 500 items sampled from parliamentary Knesset speeches for cross-genre generalization test. The predictions are produced by the best performing training checkpoint (DictaLM2.0 fine-tune).

\begin{table}[H]
\centering

\resizebox{\linewidth}{!}{%
  \begin{tabular}{lrrrr}
    \toprule
    \textbf{Class}          & \textbf{Precision} & \textbf{Recall} & \textbf{F1-score} & \textbf{Support} \\
    \midrule
    Conservative            & 0.33               & 0.82            & 0.47              & 17               \\
    Rightist                & 0.66               & 0.74            & 0.70              & 47               \\
    Democrat                & 0.76               & 0.69            & 0.72              & 74               \\
    Honest                  & 0.72               & 0.81            & 0.76              & 47               \\
    Capitalist              & 0.68               & 0.87            & 0.76              & 31               \\
    Ultra-orthodox          & 0.66               & 0.84            & 0.74              & 32               \\
    Socially-oriented       & 0.73               & 0.72            & 0.73              & 72               \\
    Liberal                 & 0.60               & 0.60            & 0.60              & 63               \\
    Leftist                 & 0.66               & 0.77            & 0.71              & 53               \\
    Security-oriented       & 0.78               & 0.75            & 0.77              & 48               \\
    Palestinian             & 0.91               & 0.86            & 0.88              & 69               \\
    Zionist                 & 0.76               & 0.81            & 0.78              & 42               \\
    \midrule
    \textbf{Micro avg}      & 0.70               & 0.76            & 0.73              & 595              \\
    \textbf{Macro avg}      & 0.69               & 0.77            & 0.72              & 595              \\
    \textbf{Weighted avg}   & 0.72               & 0.76            & 0.73              & 595              \\
    \textbf{Samples avg}    & 0.62               & 0.64            & 0.62              & 595              \\
    \bottomrule
  \end{tabular}}%
  \caption{Knesset sample results – DictaLM2.0 fine-tuned checkpoint}
\label{tab:knesset_generalization}
\end{table}

\section{Data Release}
The annotated data is released under cc-by-4.0 license. The data is publicly available on github at \url{https://github.com/guymorlan/hebid/}.

\section{Annotation}
All data has been annotated by two authors of the paper. The authors have not received direct compensation. The two annotators are Hebrew speaking women living in Israel.

\section{Model Sizes}

\begin{table}[H]
  \centering
  \small
  \begin{tabular}{@{}l r@{}}
    \toprule
    \textbf{Model}                              & \textbf{Parameters} \\
    \midrule
    mBERT (bert-base-multilingual-cased)        & 110 M     \\
    AlephBERT                                   & 110 M     \\
    HeRo                                        & 125 M     \\
    DictaBERT–base                              & 184 M     \\
    DictaBERT–large                             & 340 M     \\
    Gemma–2B                                    & 2 B       \\
    Gemma–9B                                    & 9 B       \\
    Qwen3–8B                                    & 8.2 B     \\
    DictaLM 2.0                                 & 7 B       \\
    \bottomrule
  \end{tabular}
  \caption{Model sizes (number of parameters) for all models used in this paper.}
  \label{tab:model_sizes}
\end{table}

\section{Preprocessing}

Sentence segmentation was performed using the Stanza package.

\clearpage

\begin{table*}[!h]
\section{External Validity}
\vspace{10pt}
\centering
\begin{tabular}{llcc}
\toprule
\textbf{CHES variable (2021)} & \textbf{Correlated with} & \textbf{Pearson r} & \textbf{p-value} \\
\midrule
\multirow{3}{*}{Overall Ideology (Left/Right)} & Rightist - Leftist & 0.941 & $<$ 0.001 \\
 & Rightist & 0.880 & 0.002 \\
 & Leftist & -0.815 & 0.007 \\
\midrule
\multirow{3}{*}{Social Values (Libertarian/Traditional)} & Conservative - Liberal & 0.817 & 0.007 \\
 & Conservative & 0.431 & 0.247 \\
 & Liberal & -0.838 & 0.005 \\
\midrule
\multirow{3}{*}{Civil Liberties vs. Law \& Order} & Conservative - Liberal & 0.905 & $<$ 0.001 \\
 & Conservative & 0.850 & 0.004 \\
 & Liberal & -0.845 & 0.004 \\
\midrule
\multirow{3}{*}{Economic Stance (Left/Right)} & Capitalist - Socially-oriented & 0.423 & 0.256 \\
 & Capitalist & 0.869 & 0.002 \\
 & Socially-oriented & 0.012 & 0.976 \\
\midrule
\multirow{3}{*}{State's Identity (Democratic vs. Jewish)} & Ultra-orthodox - Democrat & 0.708 & 0.033 \\
 & Ultra-orthodox & 0.649 & 0.059 \\
 & Democrat & -0.410 & 0.274 \\
\midrule
Salience of Reducing Corruption & Honest & 0.664 & 0.051 \\
\midrule
\multirow{2}{*}{Stance on Israeli-Palestinian Conflict} & Zionist & 0.865 & 0.003 \\
 & Palestinian & -0.602 & 0.086 \\
\midrule
\multirow{2}{*}{Position on a Palestinian State} & Zionist & 0.862 & 0.003 \\
 & Palestinian & -0.549 & 0.126 \\
\midrule
Position on Engagement with Arab World & Security-oriented & 0.710 & 0.032 \\
\bottomrule
\end{tabular}
\caption{Pearson correlation between CHES variables (2021) and identity discourse shares.}
\label{tab:ches_correlation}
\end{table*}

\clearpage

\begin{figure*}[!h]

\section{Additional Results}
\label{sec:additional_results}
    \centering
    \includegraphics[width=1\linewidth]{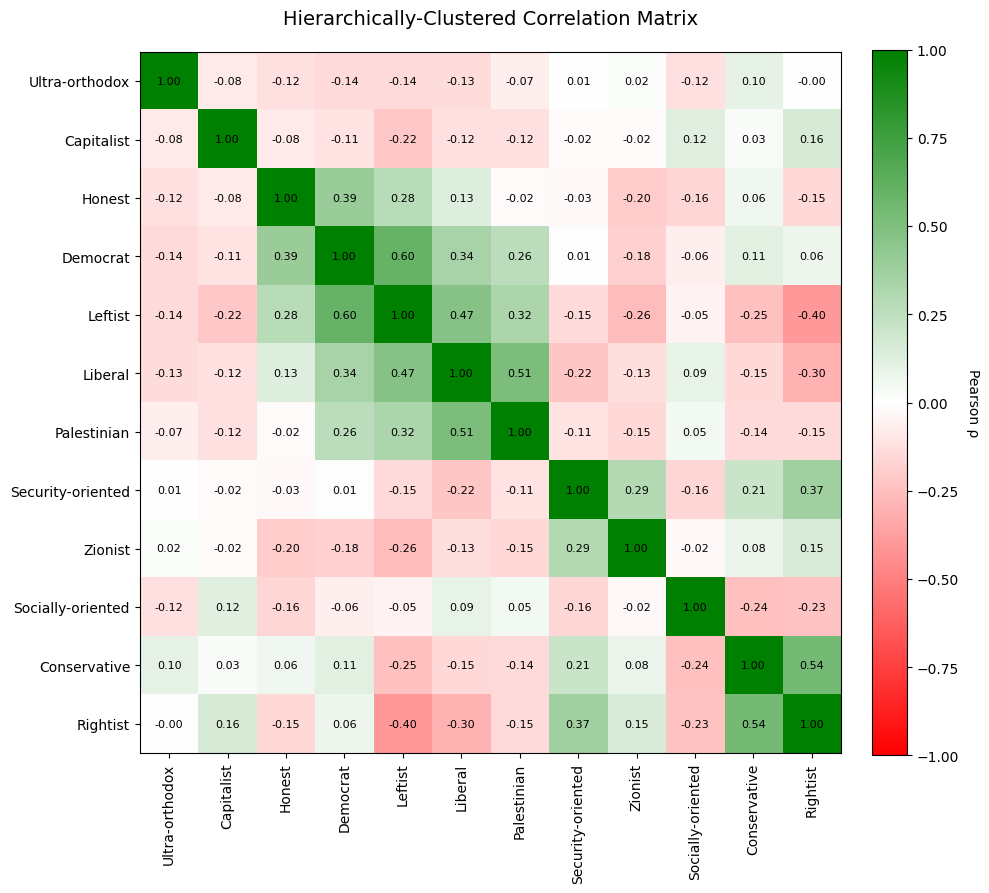}
    \caption{Identity Correlation Matrix - Facebook}
    \label{fig:enter-label}
\end{figure*}

\begin{figure*}[!h]
    \centering
    \includegraphics[width=1\linewidth]{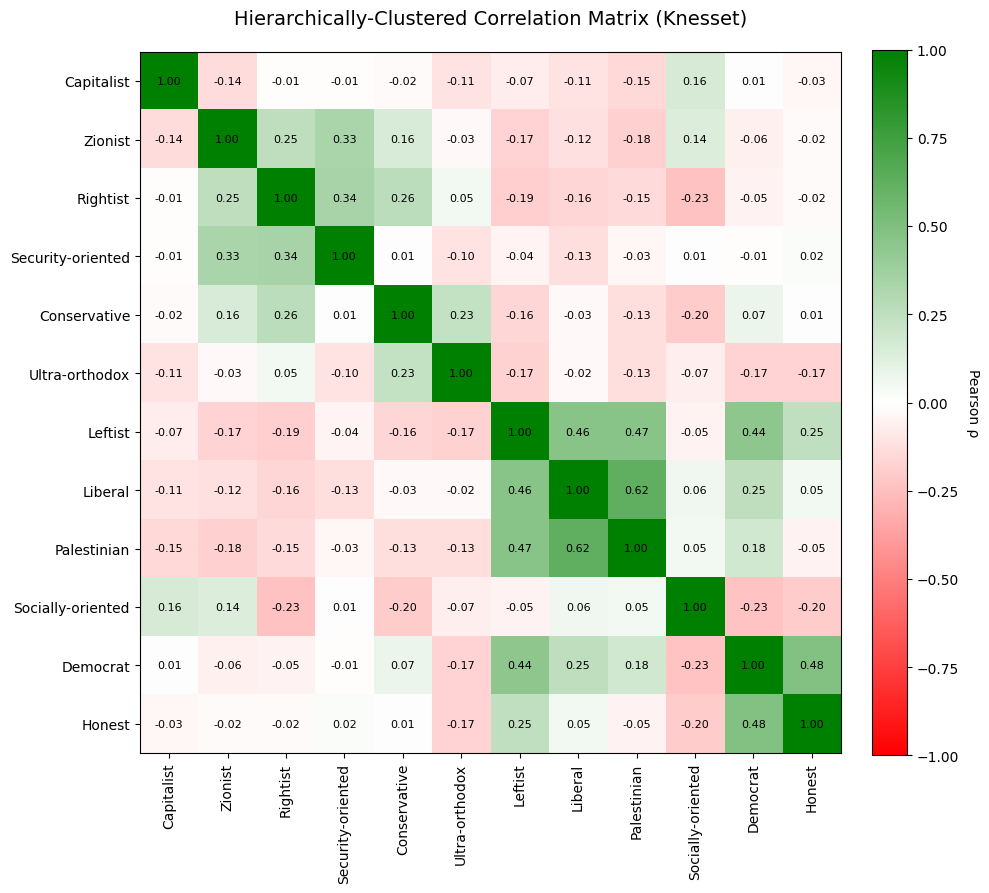}
    \caption{Identity Correlation Matrix - Knesset}
    \label{fig:cor_knesset}
\end{figure*}

\begin{figure*}[!h]
    \centering
    \includegraphics[width=1\linewidth]{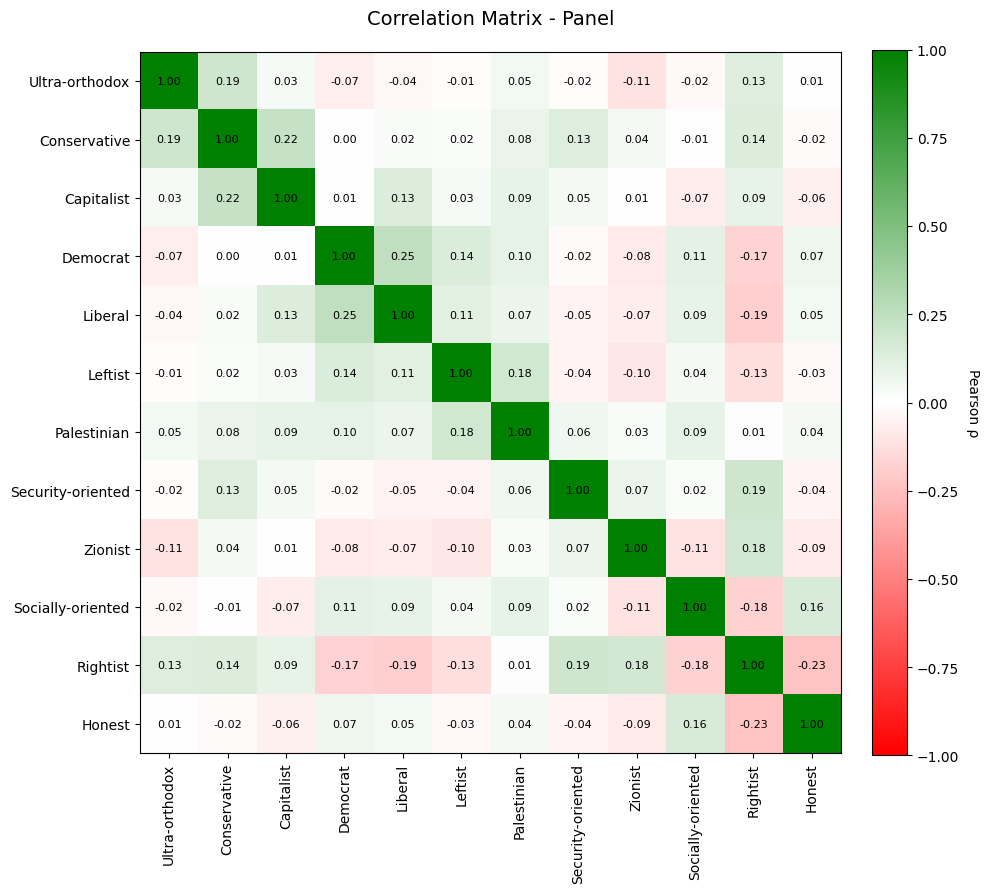}
    \caption{Identity Correlation Matrix - Survey}
    \label{fig:cor_panel}
\end{figure*}

%\begin{table}[ht]
%\centering
%\caption{Spearman correlations %between identity mention rankings}
%\label{tab:spearman_corr_long}
%\begin{tabular}{@{}l c@{}}
%\toprule
%Comparison              & %Spearman's $\rho$ \\ 
%\midrule
%Facebook vs Knesset     & 0.867    %         \\ 
%Facebook vs Survey      & 0.483             \\ 
%Knesset vs Survey       & 0.455    %         \\ 
%\bottomrule
%\end{tabular}
%\end{table}

\begin{figure*}
    \centering
    \includegraphics[width=1\linewidth]{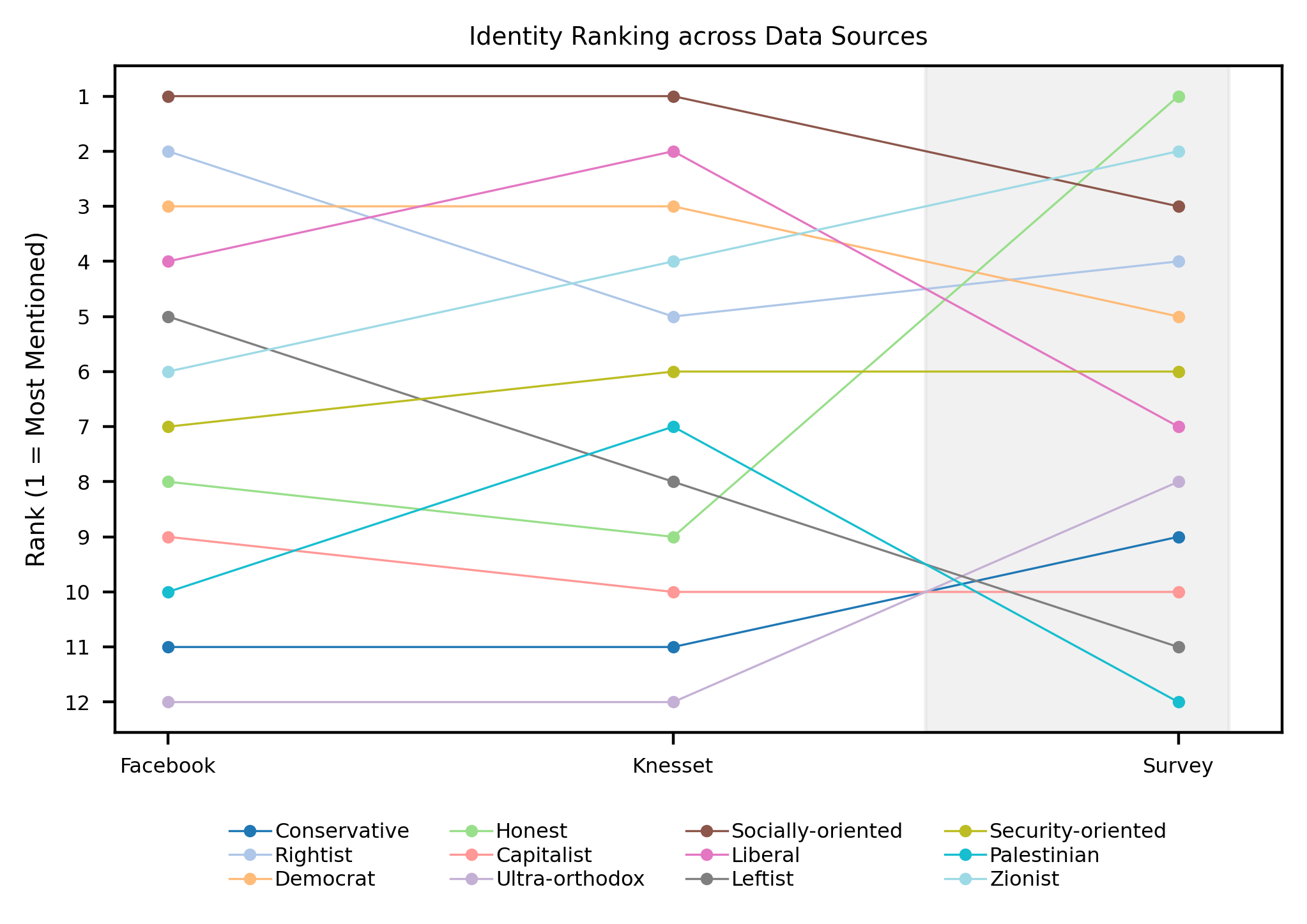}
    \caption{Identity Ranking}
    \label{fig:identity_ranking}
\end{figure*}

\begin{figure*}
    \centering
    \includegraphics[width=1\linewidth]{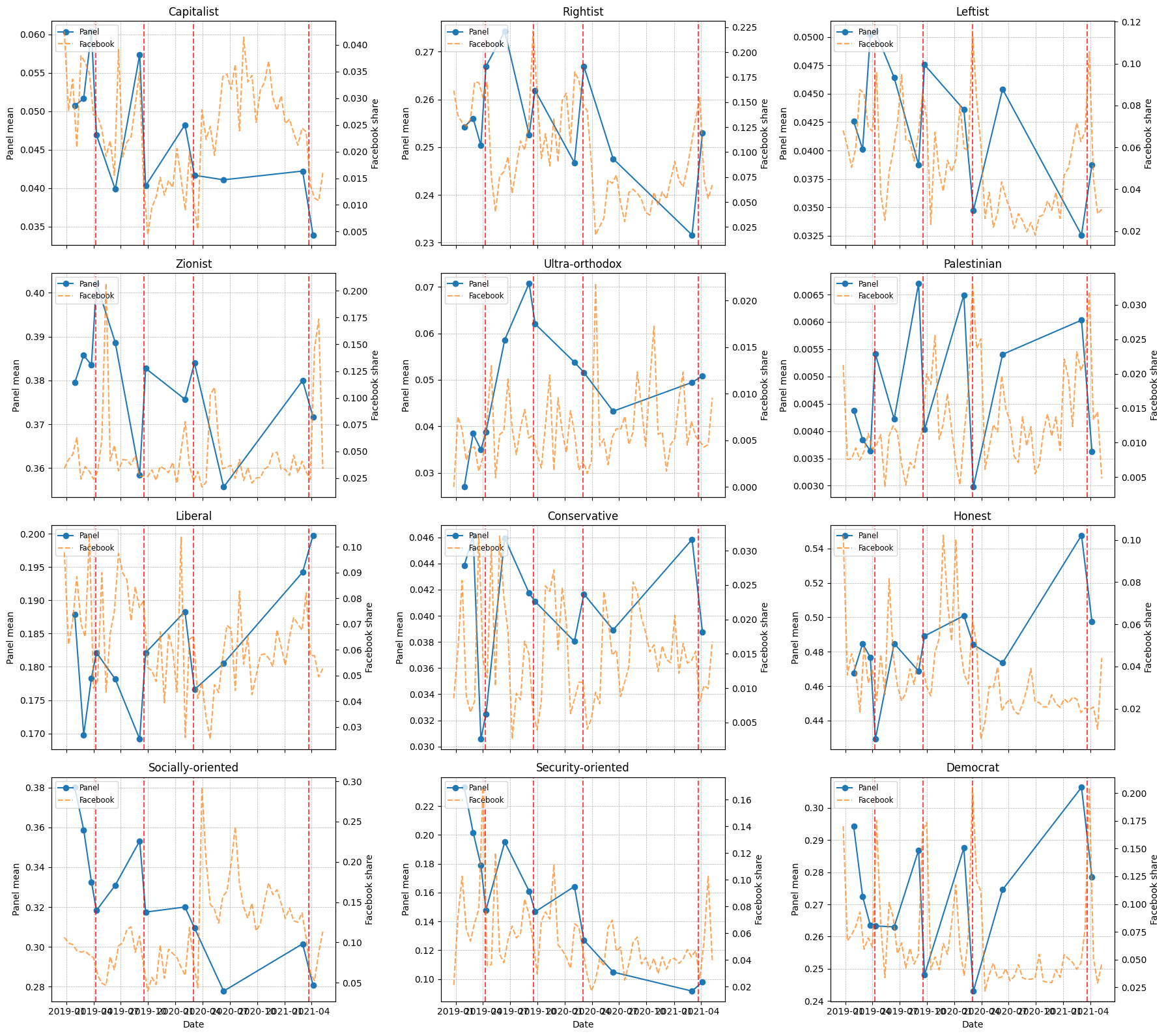}
    \caption{Per-identity time-trends on Facebook (biweekly mean) and survey}
    \label{fig:identity_timetrends}
\end{figure*}

\end{document}